\documentclass{article}
\usepackage{spconf,amsmath,epsfig}
\usepackage{url}
\usepackage{CJKnumb}
\usepackage{multirow}

\usepackage{booktabs}
\usepackage{amsfonts}
\usepackage[dvipsnames]{xcolor}

\begin{document}\sloppy

\def\x{{\mathbf x}}
\def\L{{\cal L}}

\title{Warm Diffusion: Recipe for Blur-Noise Mixture Diffusion Models}

\name{
    \textsuperscript{1}\textit{Hao-Chien Hsueh}, 
    \textsuperscript{1}\textit{Chi-En Yen}, 
    \textsuperscript{1}\textit{Wen-Hsiao Peng}, 
    \textsuperscript{1}\textit{Ching-Chun Huang}
}

\address{
    \textsuperscript{1}Department of Computer Science, \\
    National Yang Ming Chiao Tung University, Hsinchu, Taiwan \\
    E-mail: chingchun@nycu.edu.tw
}

\maketitle

\begin{abstract}
Diffusion probabilistic models have achieved remarkable success in generative tasks across diverse data types. While recent studies have explored alternative degradation processes beyond Gaussian noise, this paper bridges two key diffusion paradigms: hot diffusion, which relies entirely on noise, and cold diffusion, which uses only blurring without noise. We argue that hot diffusion fails to exploit the strong correlation between high-frequency image detail and low- frequency structures, leading to random behaviors in the early steps of generation. Conversely, while cold diffusion leverages image correlations for prediction, it neglects the role of noise (randomness) in shaping the data manifold, resulting in out-of-manifold issues and partially explaining its performance drop. To integrate both strengths, we propose Warm Diffusion, a unified Blur-Noise Mixture Diffusion Model (BNMD), to control blurring and noise jointly. Our divide-and- conquer strategy exploits the spectral dependency in images, simplifying score model estimation by disentangling the denoising and deblurring processes. We further analyze the Blur-to-Noise Ratio (BNR) using spectral analysis to investigate the trade-off between model learning dynamics and changes in the data manifold. Extensive experiments across benchmarks validate the effectiveness of our approach for image generation.
\end{abstract}

\section{Introduction}
\label{sec:intro}

Diffusion probabilistic models \cite{sohldickstein2015deepunsupervisedlearningusing, ho2020denoisingdiffusionprobabilisticmodels, nichol2021improveddenoisingdiffusionprobabilistic, song2022denoisingdiffusionimplicitmodels} have shown remarkable performance in generative tasks by learning data distributions through iterative denoising. Traditionally, these models apply Gaussian noise during the forward process and learn to reverse this via a score-based network. However, this domain-agnostic design overlooks the spectral dependency in natural images, which refers to the strong correlation between low-frequency structures and high-frequency details. This limitation reduces the efficiency of image generation.

Recent studies \cite{bansal2022colddiffusioninvertingarbitrary,daras2022softdiffusionscorematching, rissanen2023generativemodellinginverseheat, hoogeboom2024blurringdiffusionmodels, luo2023imagerestorationmeanrevertingstochastic, delbracio2024inversiondirectiterationalternative, liu2023i2sbimagetoimageschrodingerbridge, yue2023resshiftefficientdiffusionmodel, liu2024residualdenoisingdiffusionmodels} have explored alternatives to standard noise-driven diffusion, including cold diffusion methods based on deterministic transformations like blurring. While promising for specific restoration tasks, such methods often sacrifice sample diversity and generation quality.

As illustrated in Fig. 1, we propose a new Warm Diffusion process that jointly incorporates blur and noise during the forward process, aiming to balance the benefits of both hot and cold approaches. Our method introduces a controllable Blur-to-Noise Ratio (BNR) to navigate between blurring and noise, effectively leveraging spectral dependencies while preserving data manifold integrity. We analyze power spectral densities to guide BNR selection, improving training efficiency and sample realism.


\begin{figure}[t]
\centering

\includegraphics[width=0.46\textwidth]
{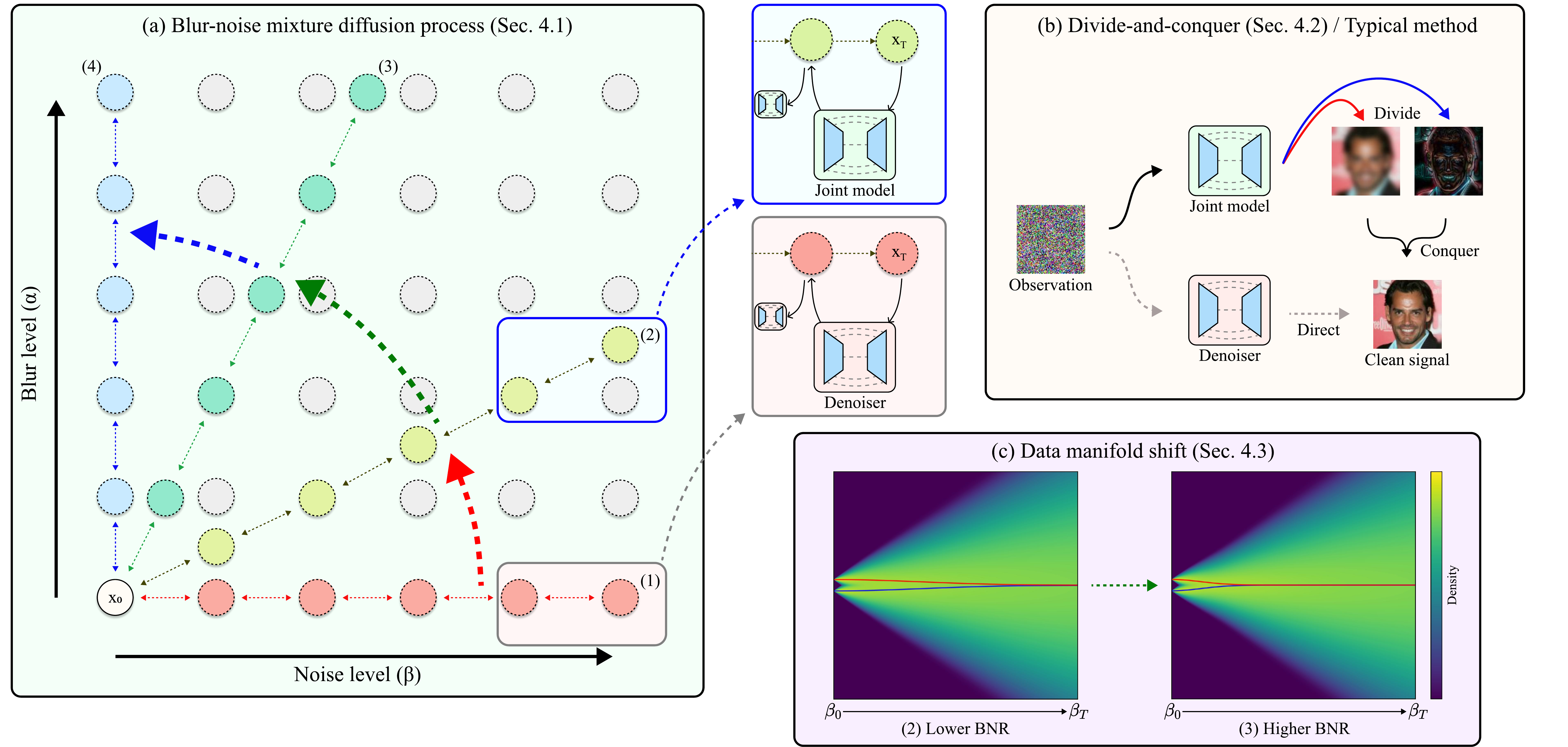}
\caption{Illustration of Warm Diffusion, the proposed two-pronged diffusion process. (a) Blur-noise mixture diffusion processes that allow flexible control over blur and noise levels, enabling a smooth transition between (1) Hot Diffusion and (4) Cold Diffusion.
(b) A divide-and-conquer strategy using a joint model for denoising and deblurring, which leverages spectral dependency to recover noise-obscured signals and restore high-frequency details.
(c) Data manifolds under different BNRs. Red and blue lines denote Gaussian means with shared low-frequency but distinct high-frequency components. Higher BNR leads to earlier merging as blurring removes high-frequency detail, potentially causing manifold shifts.}
\label{fig:teaser}
\end{figure}

Our approach, targeting image generation, has the following contributions:
\begin{itemize}
    \item \ We propose a Warm Diffusion process that combines blurring and noise in the forward process. The scheme allows flexible control over blur and noise levels, enabling joint deblurring and denoising to enhance image generation quality.
\end{itemize}
\vspace{-1.2em}
\begin{itemize}
    \item \ We introduce the new concept of Blur-to-Noise Ratio (BNR) control and show that increasing the BNR (leaning toward cold diffusion) simplifies model learning by leveraging spectral dependency effectively. However, this also increases the risk of samples deviating from the data manifold during the reverse process.
\end{itemize}
\vspace{-1.2em}
\begin{itemize}
    \item \ We select the BNR by analyzing the spectra of images and Gaussian white noise. The difference in their power spectral densities guides us to find a suitable BNR that balances two key factors: preserving the integrity of the data manifold and simplifying neural network learning.
\end{itemize}
\vspace{-1.2em}
\begin{itemize}
    \item \ Extensive experiments across various datasets show that our approach outperforms state-of-the-art diffusion methods in terms of image generation quality.
\end{itemize}

\begin{figure*}[ht]
\centering
\includegraphics[width=1\textwidth]{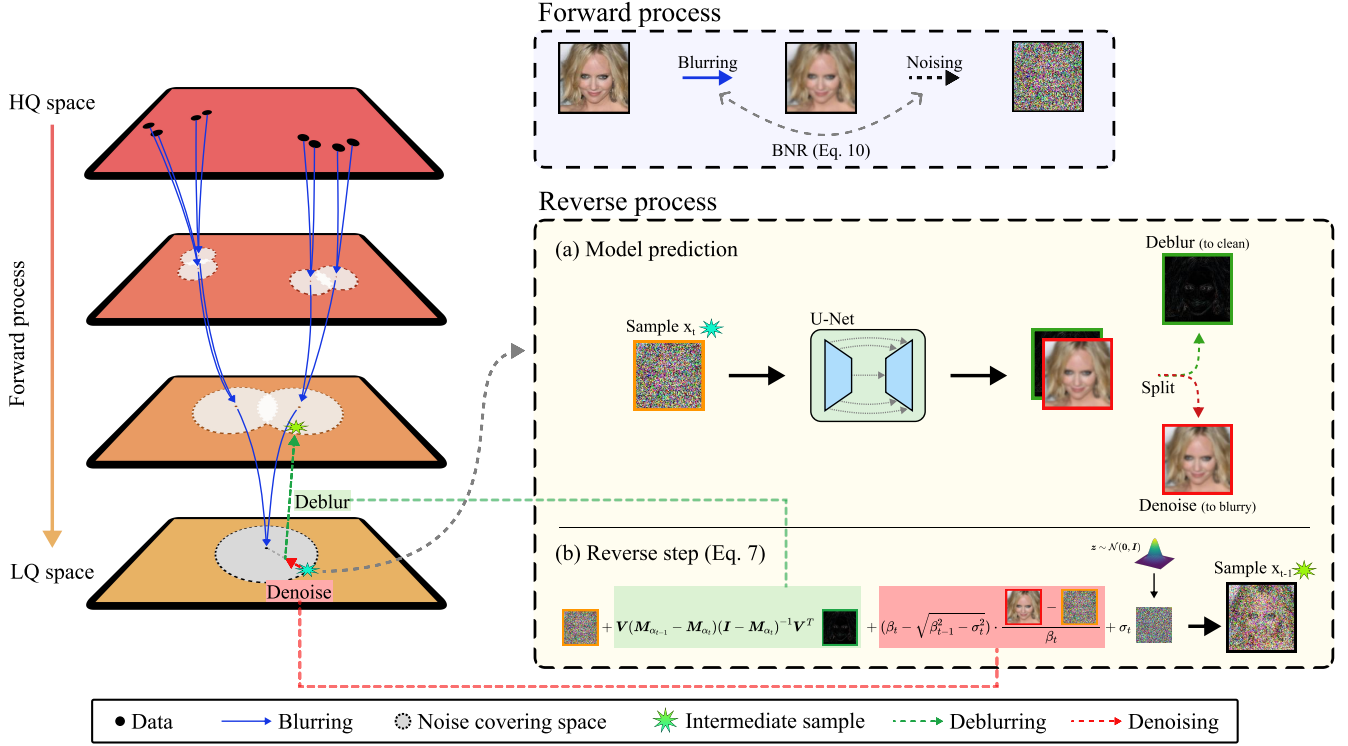}
\caption{Workflow of the proposed diffusion process. The forward process progressively applies blurring and noise, controlled by the Blur-to-Noise Ratio (BNR), to degrade the sample from high quality to low quality. During this phase, training pairs are collected to train the prediction model (e.g., U-Net) for use in the reverse process. For sample generation, the reverse process works as follows: (a) The prediction model simultaneously performs denoising and deblurring. (b) With the prediction results, the reverse step transitions the sample from step $t$ to $t-1$. Specifically, the denoiser gradually guides the sample toward a blurry prediction, while the deblurring prediction helps return the sample to a higher-quality state.}
\label{fig:workflow}
\end{figure*}


\vspace{-0.5em}
\section{Proposed Method}


To enhance image generation, we introduce a blur-noise mixture diffusion model (BNMD). BNMD extends the state space of the diffusion process to two dimensions, controlled by the corruption factors $\alpha$ and $\beta$, which represent the blur and noise levels, respectively. In the forward process, the schedules of $\alpha$ and $\beta$ determine a blend of blurring and noising operations. Consequently, the reverse process iteratively recovers a high-quality image by deblurring and denoising a sampled image drawn from a prior distribution (i.e., a standard normal distribution). As shown in ~\ref{fig:workflow}, mixing deblurring and denoising throughout the iterative image generation process distinguishes our approach from most existing diffusion models, which typically generate images through denoising alone. Moreover, our forward and generative processes are defined similarly to DDIM \cite{song2022denoisingdiffusionimplicitmodels}, but with a focus on incorporating blur-noise operations into both stages. For brevity, we primarily address the key terms from DDIM \cite{song2022denoisingdiffusionimplicitmodels} that require adaptation for our model.

\subsection{Blur-Noise Forward Diffusion Processes}\label{2D_diff}

Our blur-noise forward process has the marginal distributions for $t \in \{1, \dots, T\}$ as:
\begin{equation}
\begin{aligned}
q(\boldsymbol{x}_{\alpha_{t}, \beta_{t}}|\boldsymbol{x}_{0})=\mathcal{N}(\boldsymbol{V}\boldsymbol{M}_{\alpha_{t}}\boldsymbol{V}^{T}\boldsymbol{x}_{0},\beta_t^{2} \boldsymbol{I}),  
\end{aligned}
\label{eq.forward}
\end{equation}

where $\boldsymbol{V}^{T}$ and $\boldsymbol{V}$ denote the forward and inverse Discrete Cosine Transform (DCT), respectively. $\boldsymbol{M}_{\alpha_t}$, a diagonal matrix, specifies the Gaussian blurring mask in the DCT domain, which varies according to the Gaussian blur level $\alpha_t$. The parameter ${\beta_t}$ controls the level of Gaussian noise. The corruption sequences ${\alpha_1, \dots, \alpha_T}$ and ${\beta_1, \dots, \beta_T}$ are defined as monotonically increasing sequences. As with DDIM \cite{song2022denoisingdiffusionimplicitmodels}, Eq.~\ref{eq.forward} requires the inference transition distributions (i.e., those used in the reverse process) $q_{\sigma}(.)$ for all $t > 1$ to be:

\begin{equation}
\begin{aligned}
&q_{\sigma}(\boldsymbol{x}_{\alpha_{t-1}, \beta_{t-1}} \mid \boldsymbol{x}_{\alpha_{t}, \beta_{t}}, \boldsymbol{x}_{0})= \mathcal{N}(\boldsymbol{V} \boldsymbol{M}_{\alpha_{t-1}} \boldsymbol{V}^T \boldsymbol{x}_0 \\
&+ \sqrt{\beta_{t-1}^2 - \sigma_t^2} \cdot 
\frac{
\boldsymbol{x}_{\alpha_{t}, \beta_{t}} - \boldsymbol{V} \boldsymbol{M}_{\alpha_{t}} \boldsymbol{V}^T \boldsymbol{x}_0
}{\beta_t}, \sigma_t^2 \boldsymbol{I})
\end{aligned}
\label{eq.inference}
\end{equation}

\subsection{Generative Process in a Divide-and-Conquer Manner}\label{parameterization}
Our generative process is a Markovian process specified by learnable transition distributions $p_{\theta}(\boldsymbol{x}_{\alpha_{t-1}, \beta_{t-1}}|\boldsymbol{x}_{\alpha_{t}, \beta_{t}})$ for $t>1$. The training of these distributions involves maximizing a variational lower bound that requires minimizing among others the sum of the KL-divergence terms $KL(q_{\sigma}(\boldsymbol{x}_{\alpha_{t-1}, \beta_{t-1}} | \boldsymbol{x}_{\alpha_{t}, \beta_{t}}, \boldsymbol{x}_0)||p_{\theta}(\boldsymbol{x}_{\alpha_{t-1}, \beta_{t-1}}|\boldsymbol{x}_{\alpha_{t}, \beta_{t}})$ for $t>1$.

Notably, parameterizing $p_{\theta}(\boldsymbol{x}_{\alpha_{t-1}, \beta_{t-1}}|\boldsymbol{x}_{\alpha_{t}, \beta_{t}})$ is to leverage the knowledge about $q_{\sigma}(\boldsymbol{x}_{\alpha_{t-1}, \beta_{t-1}} | \boldsymbol{x}_{\alpha_{t}, \beta_{t}}, \boldsymbol{x}_0)$ and learn a mean function to predict the mean in Eq.~\ref{eq.inference} based on the blurry-noisy observation $\boldsymbol{x}_{\alpha_t, \beta_t}$. At the inference time, we do not have access to the input $\boldsymbol{x}_0$; as such, one straightforward approach to parameterizing $p_{\theta}(\boldsymbol{x}_{\alpha_{t-1}, \beta_{t-1}}|\boldsymbol{x}_{\alpha_{t}, \beta_{t}})$ is to learn a network $\hat{\boldsymbol{x}}_0 = F_\theta (\boldsymbol{x}_{\alpha_t, \beta_t},t)$ that makes a prediction of $\boldsymbol{x}_0$ from $\boldsymbol{x}_{\alpha_t, \beta_t}$, and have $p_{\theta}(\boldsymbol{x}_{\alpha_{t-1}, \beta_{t-1}}|\boldsymbol{x}_{\alpha_{t}, \beta_{t}}) = q_{\sigma}(\boldsymbol{x}_{\alpha_{t-1}, \beta_{t-1}} | \boldsymbol{x}_{\alpha_{t}, \beta_{t}}, \hat{\boldsymbol{x}}_0)$.

Instead of predicting $\boldsymbol{x}_0$ from $\boldsymbol{x}_{\alpha_t, \beta_t}$ directly with a single network, this work introduces a novel divide-and-conquer approach to parameterizing $p_{\theta}(\boldsymbol{x}_{\alpha_{t-1}, \beta_{t-1}}|\boldsymbol{x}_{\alpha_{t}, \beta_{t}})$. This is motivated by the observation that the mean in Eq.~\ref{eq.inference} can be expressed alternatively as:
\begin{equation}
\begin{split}
&\boldsymbol{x}_{\alpha_t, \beta_t} + \underbrace{ \boldsymbol{V}(\boldsymbol{M}_{{\alpha}_{t-1}}-\boldsymbol{M}_{\alpha_t})\boldsymbol{V}^{T} \boldsymbol{x}_0 }_{\text{add missing high-frequency detail}} \\
&+(\beta_t-\sqrt{{\beta_{t-1}^2}-\sigma_t^2})\cdot\underbrace{ \frac{\boldsymbol{V}\boldsymbol{M}_{\alpha_{t}}\boldsymbol{V}^T\boldsymbol{x}_0 - \boldsymbol{x}_{\alpha_{t}, \beta_{t}}}{\beta_t} }_{\text{direction pointing to blurry $\boldsymbol{x}_0$}}, 
\end{split}
\label{eq.reverse_sde}
\end{equation}

where we have additionally added and subtracted the same term $\boldsymbol{x}_{\alpha_t, \beta_t}- \boldsymbol{V}\boldsymbol{M}_{\alpha_t}\boldsymbol{V}^{T} \boldsymbol{x}_0$. This alternative expression suggests that the task of parameterizing $p_{\theta}(\boldsymbol{x}_{\alpha_{t-1}, \beta_{t-1}}|\boldsymbol{x}_{\alpha_{t}, \beta_{t}})$ by learning a mean function to predict that of Eq.~\ref{eq.inference} can be decomposed into two sub-tasks: deblurring and denoising. The former aims to reconstruct the high-frequency detail of $\boldsymbol{x}_0$ via the iterative generation process, while the latter is to recover a blurry version (i.e., $\boldsymbol{V}\boldsymbol{M}_{\alpha_{t}}\boldsymbol{V}^T\boldsymbol{x}_0$) of $\boldsymbol{x}_0$ from its noisy observation (i.e., $\boldsymbol{x}_{\alpha_t, \beta_t}$) by focusing on the reconstruction of the low-frequency components of $\boldsymbol{x}_0$. This interpretation allows us to learn two specialized networks for addressing these separate sub-tasks, leading to the more efficient and accurate generation of output images.

Specifically, we learn a network $D_{\theta}$ that takes the blurry-noisy observation $\boldsymbol{x}_{\alpha_{t}, \beta_{t}}=\boldsymbol{V}\boldsymbol{M}_{\alpha_{t}}\boldsymbol{V}^{T}\boldsymbol{x}_{0}+\beta_{t}\boldsymbol{\epsilon},\boldsymbol{\epsilon} \sim \mathcal{N}(\boldsymbol{0},\boldsymbol{I})$ as input to predict the noise-free yet blurry representation $\boldsymbol{V}\boldsymbol{M}_{\alpha_{t}}\boldsymbol{V}^{T}\boldsymbol{x}_{0}$ of $\boldsymbol{x}_0$. When learned successfully, $D_{\theta}$ is able to denoise $\boldsymbol{x}_{\alpha_{t}, \beta_{t}}$. For deblurring, a separate network $R_{\theta}$, which takes the same $\boldsymbol{x}_{\alpha_{t}, \beta_{t}}$ as input, is learned to update $\boldsymbol{V}\boldsymbol{M}_{\alpha_{t}}\boldsymbol{V}^{T}\boldsymbol{x}_{0}$ as $\boldsymbol{x}_0$. That is, $R_{\theta}$ aims to recover the missing high-frequency detail in $\boldsymbol{V}\boldsymbol{M}_{\alpha_{t}}\boldsymbol{V}^{T}\boldsymbol{x}_{0}$, in order to reconstruct $\boldsymbol{x}_0$. In symbols, $R_\theta$ is meant to predict $\boldsymbol{x}_{res_t} = \boldsymbol{x}_0 - \boldsymbol{V}\boldsymbol{M}_{\alpha_{t}}\boldsymbol{V}^{T}\boldsymbol{x}_{0} = \boldsymbol{V}(\boldsymbol{I} -\boldsymbol{M}_{\alpha_{t}})\boldsymbol{V}^{T}\boldsymbol{x}_{0}$. With our proposed parameterization, and given that $\boldsymbol{M}_{\alpha_t}$ is a diagonal matrix, the mean function of $p_{\theta}(\boldsymbol{x}_{\alpha_{t-1}, \beta_{t-1}}|\boldsymbol{x}_{\alpha_{t}, \beta_{t}})$ is:
\begin{equation}
\begin{aligned}
&\boldsymbol{V}(\boldsymbol{M}_{{\alpha}_{t-1}}-\boldsymbol{M}_{\alpha_{t}}){(\boldsymbol{I}-\boldsymbol{M}_{\alpha_{t}})}^{-1}\boldsymbol{V}^{T} R_\theta(\boldsymbol{x}_{\alpha_{t}, \beta_{t}},\alpha_{t}, \beta_{t}) \\ 
&+ (\beta_t-\sqrt{{\beta_{t-1}^2}-\sigma_t^2}) \cdot \frac{D_\theta( \boldsymbol{x}_{\alpha_{t}, \beta_{t}},\alpha_{t}, \beta_{t}) - \boldsymbol{x}_{\alpha_{t}, \beta_{t}}}{\beta_t} \\
&+\boldsymbol{x}_{\alpha_{t}, \beta_{t}}.
\end{aligned}
\label{eq.reverse_residual}
\end{equation} 

\vspace{1em}

Considering the equation $\boldsymbol{x}_{res_t} = \boldsymbol{V}(\boldsymbol{I} - \boldsymbol{M}_{\alpha_{t}})\boldsymbol{V}^{T}\boldsymbol{x}_{0}$, the second term $\boldsymbol{V}(\boldsymbol{M}_{\alpha_{t-1}} - \boldsymbol{M}_{\alpha_t})\boldsymbol{V}^{T} \boldsymbol{x}_0$ in Eq.~\ref{eq.reverse_sde} can be rewritten as $\boldsymbol{V} (\boldsymbol{M}_{\alpha_{t-1}} - \boldsymbol{M}_{\alpha_t}) {(\boldsymbol{I} - \boldsymbol{M}_{\alpha_t})}^{-1} \boldsymbol{V}^{T} \boldsymbol{x}_{res_t}$ to match the form of its counterpart in Eq.~\ref{eq.reverse_residual}. To approximate Eq.~\ref{eq.reverse_sde} using Eq.~\ref{eq.reverse_residual}, we then train $D_{\theta}$ and $R_{\theta}$ by minimizing the following mean square error (MSE) losses
\begin{equation}
L(D_{\theta})=\lVert D_\theta( \boldsymbol{x}_{\alpha_{t}, \beta_{t}},\alpha_{t}, \beta_{t})-(\boldsymbol{V}\boldsymbol{M}_{\alpha_{t}}\boldsymbol{V}^{T}\boldsymbol{x}_{0}) \rVert_2^2, 
\label{eq.loss_eps}
\end{equation}
\begin{equation}
L(R_{\theta})= \lVert R_\theta(\boldsymbol{x}_{\alpha_{t}, \beta_{t}},\alpha_{t}, \beta_{t}) - \boldsymbol{x}_{res_t} \rVert_2^2 .
\label{eq.loss_res}
\end{equation}

Here, given a noisy and blurry observation $\boldsymbol{x}_{\alpha_{t}, \beta_{t}} = \boldsymbol{V}\boldsymbol{M}_{\alpha_{t}}\boldsymbol{V}^{T}\boldsymbol{x}_{0} + {\beta_{t} \boldsymbol{\epsilon}}$ at time $t$, along with the corresponding corruption factors, $\alpha_{t}$ and $\beta_{t}$, $D_{\theta}$ (Denoiser) is trained to recover the pure blurry image by removing the noise component, while $R_{\theta}$ (Deblurrer) extracts the high-frequency detail $\boldsymbol{x}_{res_t}$. Finally, at time $t$, the prediction of the clean image is given by $\Hat{\boldsymbol{x}}_{0} = F_{\theta}=D_{\theta} + R_{\theta}$.

\begin{figure*}[h]
\centering
\includegraphics[width=1.0\textwidth]{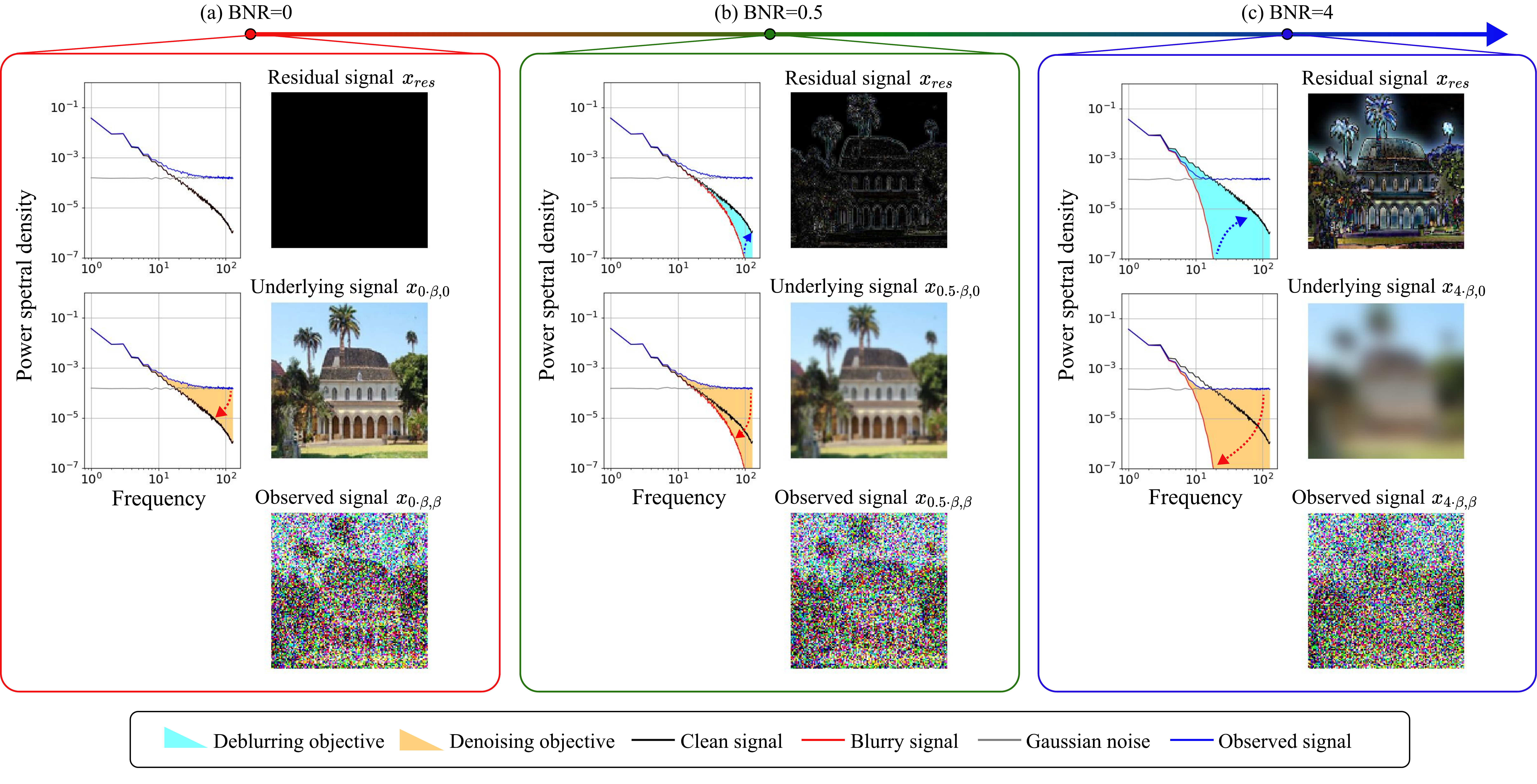}
\caption{
Impact of Varying BNRs on Model Behavior. We illustrate the observed signal, denoising target, and deblurring target, along with their respective signal spectrum analyses, across different BNR values. From left to right, the noise level remains constant while the BNR value increases. As the BNR rises, the denoising task (red arrow) becomes progressively easier, shifting more responsibility to the deblurring task (blue arrow) and effectively utilizing the spectral dependency of images. In contrast, when BNR = 0, the model requires a stronger denoiser to directly generate the image, without leveraging the spectral dependency assistance from the deblurrer.
}
\label{fig:BNR_target}
\end{figure*}

\subsection{The Impact of Blur-to-noise Ratio on Model Behavior and Data Manifold}\label{BNR}
We have demonstrated how to combine two degradation factors---blurring and noising---into a unified diffusion process and introduced training objectives that simultaneously address deblurring and denoising for image generation. However, the relationship between the blur and noise levels remains unclear. To explore this connection, we define a factor called the Blur-to-Noise Ratio (BNR):
\begin{equation}
BNR=\frac{Blur\ Level}{Noise\ Level}=\frac{\alpha}{\beta},
\label{eq.BNR} 
\end{equation}
which represents the ratio of the blur level to the noise level. As illustrated in Fig.~\ref{fig:teaser}, increasing the BNR value from 0 to $\infty$ transitions the diffusion path from hot to cold diffusion. To further investigate the model behavior with varying BNR values, we examine the learning objectives introduced in Sec.~\ref{parameterization}, which consist of two distinct branches targeting denoising and deblurring, respectively. Fig.~\ref{fig:BNR_target} shows the signal spectra of images and the corresponding training objectives for different BNR values. With a constant noise level, an increase in BNR would raise the blur level, thereby simplifying the denoising task. This shift occurs because the denoiser no longer needs to restore high-frequency detail, transferring that responsibility to the deblurring task. Additionally, by effectively utilizing spectral dependency, the deblurrer can efficiently learn a mapping function from the low-frequency observation to its high-frequency counterpart.

\begin{figure*}[t]
\centering
\includegraphics[width=1.0\textwidth]{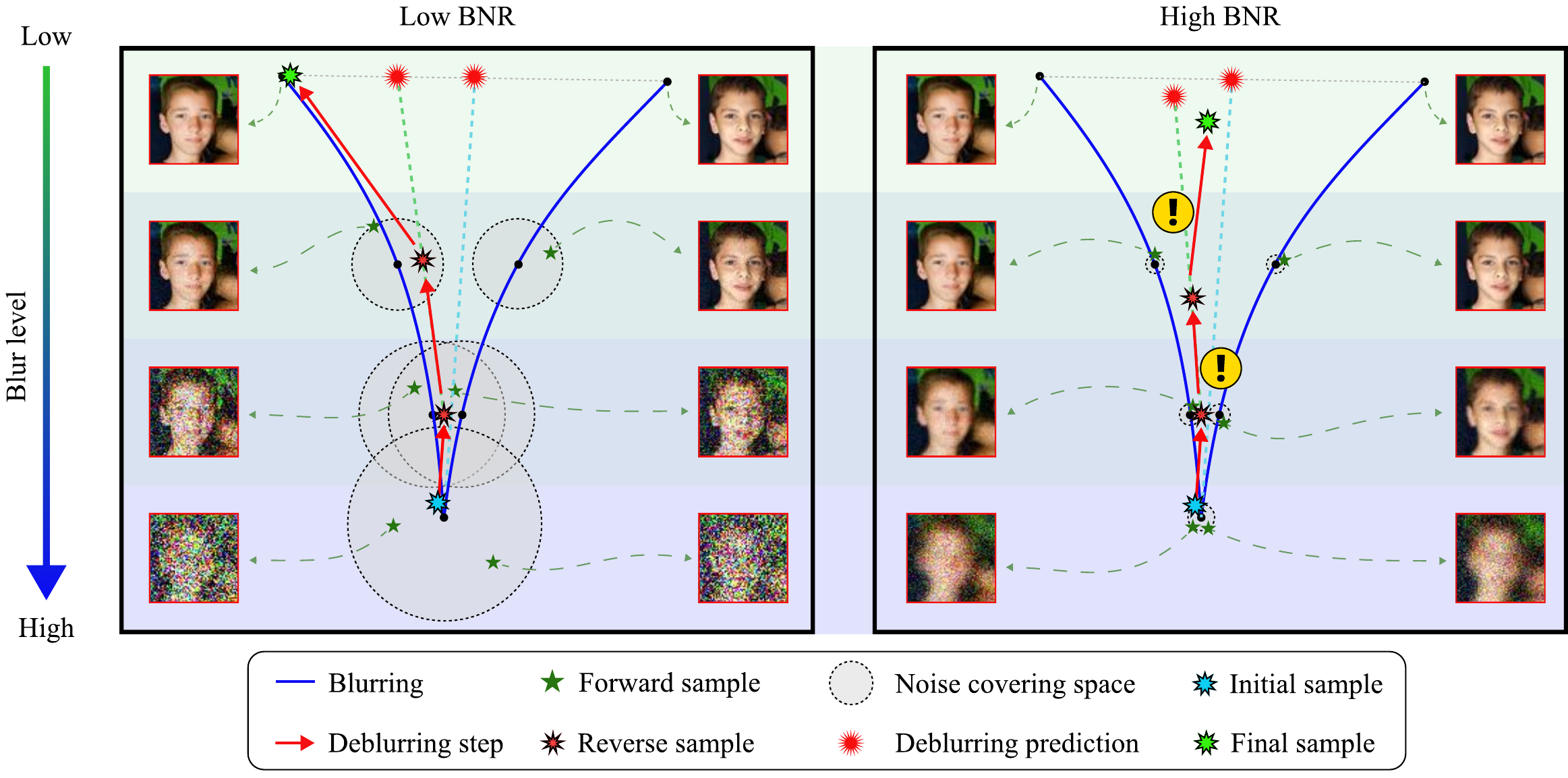}
\caption{
Illustration of the connection between BNR and the data manifold. When comparing two different BNRs at the same blur level, a higher BNR corresponds to a smaller noise scale, resulting in a narrower noise-covering space, as shown on the right. In the deblurring (reverse) step, a sample is guided toward the deblurring target, representing the mean image of all possible paired outputs. It is important to note that during the forward process, a single low-quality (LQ) sample is typically paired with multiple high-quality (HQ) samples for training. Due to this ill-posed nature of the deblurring task, samples with higher BNR values are more likely to deviate from the data manifold during the transition. Once samples fall out of the manifold, the neural network struggles to produce accurate predictions, leading to a decline in generation quality.
}
\label{fig:data_manifold}
\end{figure*}

In addition to affecting model behavior, varying BNR values also lead to shifts in data manifolds during forward iterations. To explore this in greater detail, we compare the data manifolds under low and high BNR scenarios with the same blur level in Fig.~\ref{fig:data_manifold}. At each forward or reverse step, low BNR cases exhibit a larger noise-covering space due to higher randomness compared to high BNR cases. Consequently, a high BNR results in reduced data diversity, making generated samples more likely to fall out of the data manifold, particularly during the early forward steps. When encountering out-of-manifold cases in sample generation, the diffusion network must handle degraded samples that were not seen during training, leading to less predictable outcomes and lower generation quality.

The analyses above reveal an inherent trade-off between model learning dynamics and data manifold shifts, where the choice of BNR value plays a crucial role in generation quality. As the diffusion process transitions from hot to cold, the model increasingly depends on leveraging the spectral dependency of images for learning. However, this shift also introduces the risk of divergence from the data manifold during generation, potentially resulting in degraded performance.

\subsection{Selecting BNR}\label{Policy}
We've discussed how BNR influences model behavior and the data manifold. A higher BNR allows the model to better exploit spectral dependency, simplifying learning by shifting the focus to deblurring. However, it also heightens the risk of samples deviating from the data manifold. Striking the right balance is crucial, as one must weigh the benefits of more straightforward neural network training against the risk of such deviations. This raises the question of whether a better BNR value exists that preserves the data manifold's integrity while enhancing model training.

An interesting observation from previous studies (\cite{VANDERSCHAAF19962759, rissanen2023generativemodellinginverseheat}) is that the power spectral density of natural images follows an approximate power law, ${1}/{f^\alpha}$, where $\alpha \approx 2$. In contrast, Gaussian white noise exhibits a flat frequency response across all frequency bands. This discrepancy results in a much lower signal-to-noise ratio (SNR) in high-frequency bands compared to low-frequency ones. When the SNR in these bands is sufficiently low, the observed signal becomes dominated by noise, helping to maintain the integrity of the data manifold while attenuating the image signal in these bands.

As shown in Fig.~\ref{fig:BNR_target}(b), our empirical findings indicate that selecting \textbf{BNR=0.5} causes the image signal to begin attenuating when noise intensity exceeds the signal in these frequency bands, keeping the blurry-noisy signal comparable to the noisy signal in hot diffusion. Beyond this threshold, Fig.~\ref{fig:BNR_target}(c) illustrates that for a higher BNR value, the observed signal diverges from that in hot diffusion, as the image signal attenuates significantly before noise dominates those frequency bands.

\section{Experiments}
\subsection{Image Generation}
\noindent\textbf{Datasets.} 
We evaluate our method on CIFAR-10 \cite{cifar10} $32 \times 32$, FFHQ \cite{ffhq} $64 \times 64$, and LSUN-church $128 \times 128$, covering a range of semantics and resolutions. CIFAR-10 \cite{cifar10} is used for both unconditional and class-conditional generation. FFHQ \cite{ffhq} offers face images with a higher degree of shared structure compared to general scenes. LSUN-church contains complex layouts, enabling validation at higher resolutions.



\noindent\textbf{Implementation Details.} 
We adopt DDPM++/NCSN++ architectures, training strategies, and hyperparameters from EDM. To support blur and noise conditioning, we modify the network to take two inputs and double its output channels for deblurring and denoising, respectively. Most components are shared between the two branches, so varying BNR does not change model capacity. Sampling follows EDM's Heun's 2\textsuperscript{nd}-order solver.

\begin{table}[h]
\centering
\small
\caption{Quantitative results and comparison for $32 \times 32$ and $64 \times 64$ image generation tasks on CIFAR-10 \cite{cifar10} and FFHQ \cite{ffhq} datasets correspondingly. Lower FID and higher IS scores indicate better sample quality. NFE denotes the \textbf{``Number of Function Evaluations''}. The best results are highlighted in bold; the second-best results are underlined.}
\label{tab:cifar10}
\scalebox{0.75}{
\begin{tabular}{l|ccc}
\toprule
Methods & NFE $\downarrow$ & FID $\downarrow$ & IS $\uparrow$\\
\midrule  
\multicolumn{4}{c}{Unconditional CIFAR-10} \\
\midrule  
Cold Diffusion (Blur) \cite{bansal2022colddiffusioninvertingarbitrary} & 50 & 80.08 & - \\
IHDM \cite{rissanen2023generativemodellinginverseheat} & 200 & 18.96 & - \\
Blurring Diffusion \cite{hoogeboom2024blurringdiffusionmodels} & 1000 & 3.17 & 9.51 \\
EDM \cite{karras2022elucidatingdesignspacediffusionbased} & 35 & 1.97 & 9.78 \\
EDM-ES \cite{ning2024elucidatingexposurebiasdiffusion} & 35 & 1.95 & - \\
STF \cite{xu2023stabletargetfieldreduced} & 35 & 1.92 & \underline{9.79} \\
PFGM++ \cite{xu2023pfgmunlockingpotentialphysicsinspired} & 35 & \underline{1.91} & - \\
\midrule  
Ours & 35 & \textbf{1.85} & \textbf{10.02} \\
\midrule  
\multicolumn{4}{c}{Class-conditional CIFAR-10} \\
\midrule  
EDM \cite{karras2022elucidatingdesignspacediffusionbased} & 35 & 1.79 & - \\
EDM-ES \cite{ning2024elucidatingexposurebiasdiffusion} & 35 & 1.80 & - \\
PFGM++ \cite{xu2023pfgmunlockingpotentialphysicsinspired} & 35 & \underline{1.74} & -  \\
\midrule  
Ours & 35 & \textbf{1.68} & \textbf{10.19} \\
\midrule  
\multicolumn{4}{c}{FFHQ $64\times64$} \\
\midrule  
EDM \cite{karras2022elucidatingdesignspacediffusionbased} & 79 & 2.53 & - \\
PFGM++ \cite{xu2023pfgmunlockingpotentialphysicsinspired} & 79 & \underline{2.43} & - \\
\midrule  
Ours & 79 & \textbf{2.29} & \textbf{3.41} \\
\bottomrule
\end{tabular}
}
\end{table}

\begin{table}[h]
\centering
\small
\caption{Quantitative results and comparisons for $128 \times 128$ image generation tasks on the unconditional LSUN-church \cite{lsun} dataset. For a fair comparison, we evaluate sample quality using the same number of samples as in previous studies.}
\label{tab:church}
\scalebox{0.8}{
\begin{tabular}{l|cc}
\toprule
Methods & NFE $\downarrow$ & FID $\downarrow$ \\ 
\midrule  
\multicolumn{3}{l}{Number of samples = 10k} \\
\midrule  
Denoising Diffusion \cite{hoogeboom2024blurringdiffusionmodels} & 1000 & 4.68 \\
Blurring Diffusion \cite{hoogeboom2024blurringdiffusionmodels} & 1000 & 3.88 \\
Ours  & 511 & 3.47 \\
\midrule  
\multicolumn{3}{l}{Number of samples = 50k} \\
\midrule  
IHDM \cite{rissanen2023generativemodellinginverseheat} & 400 & 45.06 \\
Ours & 511 & 2.56 \\
\bottomrule
\end{tabular}
}
\end{table}

\begin{table}[h]
\centering
\small
\caption{Ablation study on the impact of different BNR values for CIFAR-10, with a fixed number of sampling steps (NFE=35).}
\label{tab:BNR_ablation}
\scalebox{0.85}{
\begin{tabular}{l|cc}
\toprule
BNR  & FID $\downarrow$ & IS $\uparrow$ \\
\midrule
0 (EDM \cite{karras2022elucidatingdesignspacediffusionbased})  & 1.97  & 9.78 \\
0.1 & 1.97  & 9.96 \\
0.3 & 1.90  & 10.02 \\
0.5 & 1.85  & 10.02 \\
0.65 & 1.91  & 10.00 \\
1 & 2.01  & 9.96 \\
2 & 2.57  & 9.89 \\
10 & 11.97  & 8.51 \\
\bottomrule
\end{tabular}
}
\end{table}

\noindent\textbf{Performance Comparison.}
We evaluate generation quality using Fr\'echet Inception Distance (FID) and Inception Score (IS). Following standard practice, we sample 50K images over three rounds and report the best scores. As shown in Tab.~\ref{tab:cifar10} and Tab.~\ref{tab:church}, our method consistently improves over EDM on CIFAR-10 \cite{cifar10} (both conditional and unconditional) and FFHQ \cite{ffhq}, with better FID, IS, and fewer sampling steps. We also outperform Cold Diffusion, IHDM, and Blurring Diffusion on LSUN-church $128 \times 128$, demonstrating stronger performance on complex high-resolution data. To ensure fairness, we fix the number of samples for FID evaluation.

\begin{figure}[!t]
\centering
\begin{minipage}[t]{0.48\linewidth}
\centering
\includegraphics[width=\linewidth]{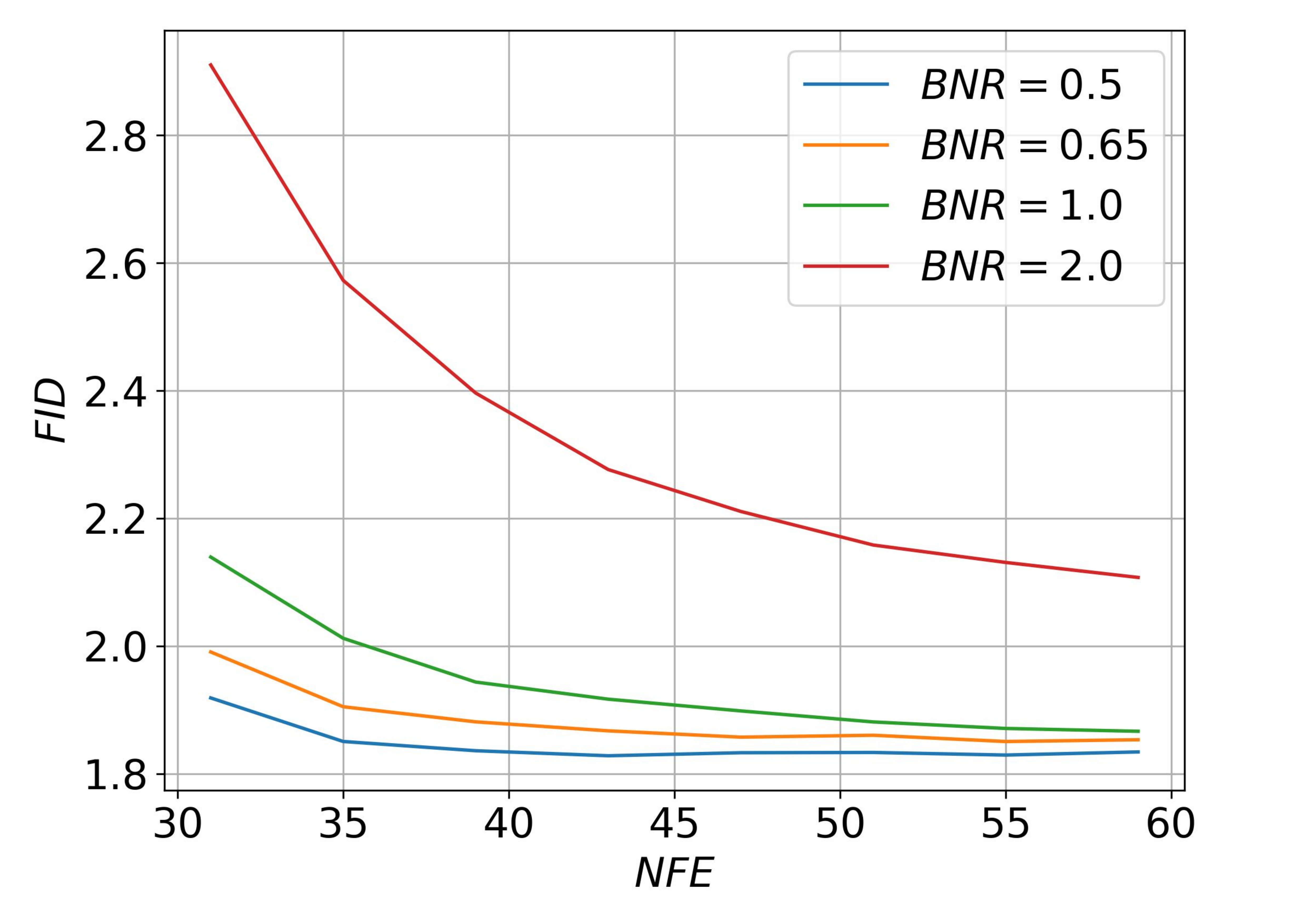}
\caption{Illustration of sample quality corresponding to different BNR and NFE. Each curve represents a specific BNR value. As shown in the chart, higher BNR values require more sampling steps to achieve better sample quality.}
\label{fig:BNR_vs_NFE}
\end{minipage}\hfill
\begin{minipage}[t]{0.48\linewidth}
\centering
\includegraphics[width=\linewidth]{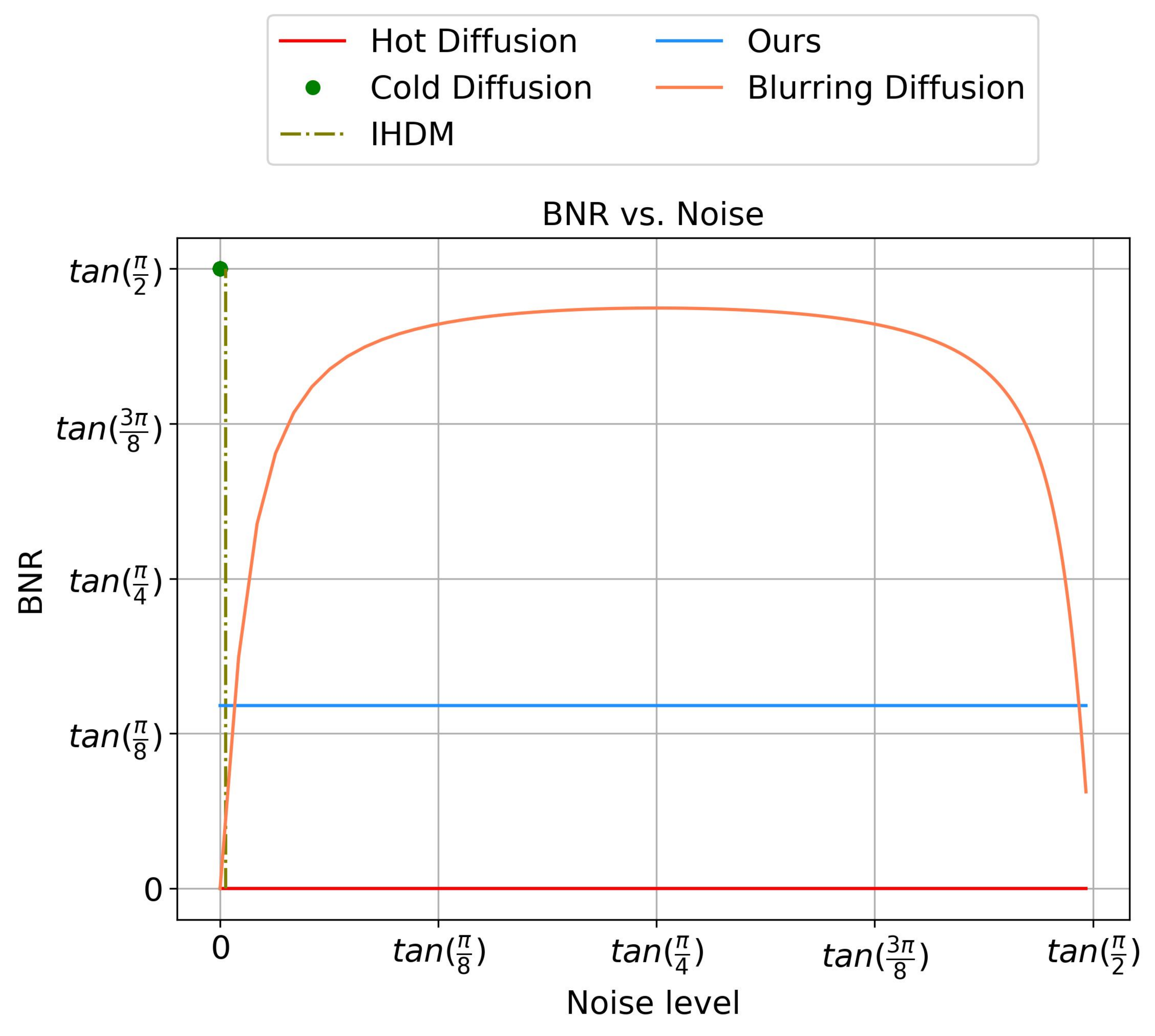}
\caption{Comparison of blur and noise schedules with previous methods. Most related studies, except the hot diffusion, selected higher BNR values in their schedules, leading to out-of-manifold issues.}

\label{fig:schedule_prev}
\end{minipage}
\end{figure}

\subsection{Analysis of Different BNRs and Scheduling}
\noindent\textbf{Relation between BNRs and Data Manifolds.}
We validate our assumptions by testing BNR values from 0 to 10. As shown in Tab.~\ref{tab:BNR_ablation}, sample quality improves with increasing BNR up to 0.5 using the same sampling steps as EDM \cite{karras2022elucidatingdesignspacediffusionbased}, but drops sharply beyond that, eventually underperforming the baseline. This supports our hypothesis in Sec.~\ref{Policy} that BNR = 0.5 achieves a balance between limiting manifold shifts and enhancing learning. As discussed in Sec.~\ref{Policy}, larger BNR values cause greater manifold shifts, making samples more likely to fall outside and degrade quality. Fig.~\ref{fig:BNR_vs_NFE} further shows that higher BNR values require more sampling steps to recover generation quality, aligning with our analysis.

\begin{table}[h]
\centering
\small
\caption{
We re-implement Blurring Diffusion using our parameterization and training scheme on CIFAR-10. Results marked with $^{\ast}$ are those reported by \cite{hoogeboom2024blurringdiffusionmodels}.
}
\label{tab:re_blurring_diff}
\begin{tabular}{l|ccc}
\toprule
BNR Schedule  & NFE $\downarrow$ & FID $\downarrow$ & IS $\uparrow$ \\
\midrule
\multirow{7}{*}{Blurring Diffusion \cite{hoogeboom2024blurringdiffusionmodels}} & 35 & 12.97 & 8.57 \\
 & 79 & 4.13 & 9.27 \\
 & 159 & 2.91 & 9.46 \\
 & 239 & 2.77 & 9.52 \\
 & 319 & 2.73 & 9.54 \\
 & 399 & 2.71 & 9.54 \\
 & 999 & 2.68 & 9.56 \\
\midrule
Blurring Diffusion$^{\ast}$ \cite{hoogeboom2024blurringdiffusionmodels} & 1000 & 3.17 & 9.51 \\
\bottomrule
\end{tabular}
\end{table}

\begin{table}[!t]
\centering
\small
\caption{An ablation study on different parameterization of $\boldsymbol{x}_0$ and $\boldsymbol{V}\boldsymbol{M}_{\alpha_t}\boldsymbol{V}^T \boldsymbol{x}_0$ in Eq.~\ref{eq.reverse_sde}. We fix a constant number of sampling steps (NFE=35) to investigate various parameterizations for the generation task using CIFAR-10.}
\label{tab:para_ablation}
\scalebox{0.76}{ 
\begin{tabular}{c|cc|cc|c}
    \toprule
    &
    \multicolumn{2}{c|}{Training Objectives} &     
    \multicolumn{2}{c|}{Parameterization of} &
     \\
    &
    $R_{\theta}$ & 
    $D_{\theta}$ & 
    $\boldsymbol{x}_0$ &
    $\boldsymbol{V}\boldsymbol{M}_{\alpha_t}\boldsymbol{V}^T \boldsymbol{x}_0$ &
    FID $\downarrow$
    \\
    \midrule
    (a) & $\boldsymbol{x}_0$ & - & $R_{\theta}$ & $\boldsymbol{V}\boldsymbol{M}_{\alpha_t }\boldsymbol{V}^TR_{\theta}$  & 13.19 \\
    (b) & - & $\boldsymbol{V}\boldsymbol{M}_{\alpha_t}\boldsymbol{V}^{T}\boldsymbol{x}_0$ & $\boldsymbol{V}{(\boldsymbol{M}_{\alpha_t})}^{-1}\boldsymbol{V}^TD_{\theta}$ & $D_{\theta}$  & 9.09 \\
    (c) & $\boldsymbol{x}_0$ & $\boldsymbol{V}\boldsymbol{M}_{\alpha_t}\boldsymbol{V}^{T}\boldsymbol{x}_0$ & $R_{\theta}$ & $D_{\theta}$  & 1.98 \\
    (d) & $\boldsymbol{x}_{res_t}$ & $\boldsymbol{V}\boldsymbol{M}_{\alpha_t}\boldsymbol{V}^{T}\boldsymbol{x}_0$ & $\boldsymbol{V}{({\boldsymbol{I}-\boldsymbol{M}_{\alpha_t}})}^{-1}\boldsymbol{V}^TR_{\theta}$ & $D_{\theta}$  & 1.85 \\
    \bottomrule
\end{tabular}
}
\end{table}

\noindent\textbf{Revisiting the BNR Scheduling in Prior Studies.}
We conduct experiments to compare the BNR schedules from prior studies with our proposed method. Specifically, we re-implement the BNR schedule from Blurring Diffusion \cite{hoogeboom2024blurringdiffusionmodels} under the same experimental conditions to eliminate the effects of differing parameterization techniques. The results, presented in Tab.~\ref{tab:re_blurring_diff}, indicate that Blurring Diffusion's BNR schedule results in poor generation quality when using fewer sampling steps, although quality improves significantly with more steps. This highlights a limitation in Blurring Diffusion's BNR scheduling, as depicted in Fig.~\ref{fig:schedule_prev}, where higher BNR values necessitate additional sampling steps to prevent the reverse process from deviating from the data manifold. Consequently, fewer steps result in poorer sample quality. These findings help explain why previous approaches, such as cool diffusion, have struggled to generate high-quality samples, particularly under limited sampling budgets.

\subsection{Ablation Studies on Training Objectives and Variations of Parameterizations}
\label{ablation_para}
In Sec.~\ref{parameterization}, we reformulate the reverse function, Eq.~\ref{eq.reverse_sde}, to simplify the neural network's training objectives through a divide-and-conquer approach. This method separates the task of predicting the clean signal $\boldsymbol{x}_0$ into two sub-tasks: denoising to a blurry signal and predicting the residual signal for deblurring. In this subsection, we explore various parameterization strategies derived from Eq.~\ref{eq.reverse_sde} and evaluate their performances, demonstrating the advantages of our divide-and-conquer strategy for model learning.

As shown in Tab.~\ref{tab:para_ablation}(a), using a single branch to directly predict the entire clean signal results in significantly poorer generation quality, as this task proves too challenging. In Tab.~\ref{tab:para_ablation}(b), shifting the target to learn a blurry signal yields a slight improvement in sample quality since this target is easier to model; however, it still faces issues with inaccuracies in high-frequency components, which may be exacerbated during sampling, occasionally resulting to noisy patterns in the generated samples. In Tab.~\ref{tab:para_ablation}(c), employing two branches to predict both clean and blurry signals effectively addresses these challenges, leading to substantially better results, though they remain comparable to those of hot diffusion models as EDM \cite{karras2022elucidatingdesignspacediffusionbased}. Finally, in Tab.~\ref{tab:para_ablation}(d), the proposed divide-and-conquer strategy further improves performance, benefiting especially from the BNR schedule and the parameterization.

\section{Conclusions}
In this paper, we introduce a unified Warm Diffusion framework that effectively bridges the gap between hot and cold diffusion models while addressing their inherent limitations. Our analysis reveals that hot diffusion models underutilize the spectral dependency of images, whereas cold diffusion models risk reverse sampling steps that deviate from the data manifolds. By examining the Blur-to-Noise Ratio (BNR), we uncover its significant influence on model behavior and data manifolds. This insight enables us to propose a strategy for balancing the trade-off between hot and cold diffusion, ultimately enhancing diffusion models for image generation. Experimental results across various benchmarks validate the effectiveness of our approach, demonstrating improvements in sample quality over state-of-the-art diffusion models.

\bibliographystyle{IEEEbib}
\bibliography{cvgip2019template}

\end{document}